\documentclass[twoside,11pt]{article}

\usepackage{jmlr2e}
\usepackage{algorithm}
\usepackage{algpseudocode}
\usepackage{hyperref}
\usepackage{amsmath}
\usepackage{amssymb}
\usepackage{lipsum}
\usepackage{graphicx}
\usepackage{multirow}
\usepackage{epstopdf}
\usepackage{caption}
\usepackage{tablefootnote}
\usepackage{subcaption}

\ShortHeadings{Classifier comparison using precision}{Gondara}
\firstpageno{1}

\begin{document}

\title{Classifier comparison using precision}

\author{\name Lovedeep Gondara \email lgondara@sfu.ca \\
       \addr Department of Computer Science\\
       Simon Fraser University\\
       Burnaby, BC, Canada
       }

\editor{}

\maketitle

\begin{abstract}
New proposed models are often compared to state-of-the-art using statistical significance testing. Literature is scarce for classifier comparison using metrics other than accuracy. We present a survey of statistical methods that can be used for classifier comparison using precision, accounting for inter-precision correlation arising from use of same dataset. Comparisons are made using per-class precision and methods presented to test global null hypothesis of an overall model comparison. Comparisons are extended to multiple multiclass classifiers and to models using cross validation or its variants. Partial Bayesian update to precision is introduced when population prevalence of a class is known. Applications to compare deep architectures are studied. 
\end{abstract}


\section{Introduction}
Classification models are often compared to test a global null hypothesis ($H_0$) of one performing significantly better than other(s). There is no standard framework, nor it is clearly defined what statistical tests to use on what performance metrics. This often results in arbitrary choices as noted by \cite{demvsar2006statistical} with classification accuracy being used most often.

Precision is an important performance statistic, especially useful in rare class predictions. It is the probability of positive prediction conditioned on classifier results and is often calculated per-class with an average reported as a point estimate
\begin{equation}\label{mainprec}
	P_C = \dfrac{1}{N} \sum_{i=1}^{N} P_{Ci}
\end{equation}
where $P_{Ci}$ is precision for class $i$ and $N$ is number of classes.

\begin{figure}
	\centering
	\begin{subfigure}{.5\textwidth}
		\includegraphics[width=.9\linewidth]{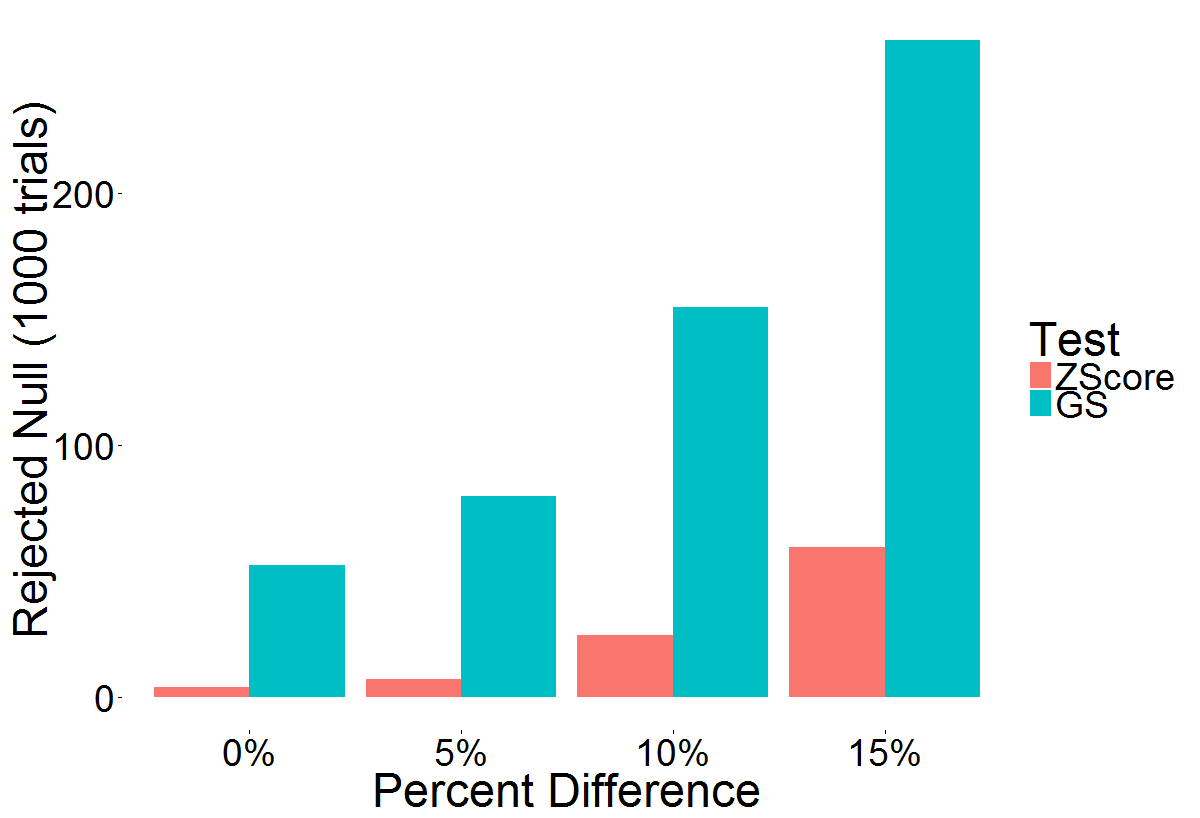}
		\caption{Sample size=100}
		\label{subfiga}
	\end{subfigure}\hfill
	\begin{subfigure}{.5\textwidth}
		\includegraphics[width=.9\linewidth]{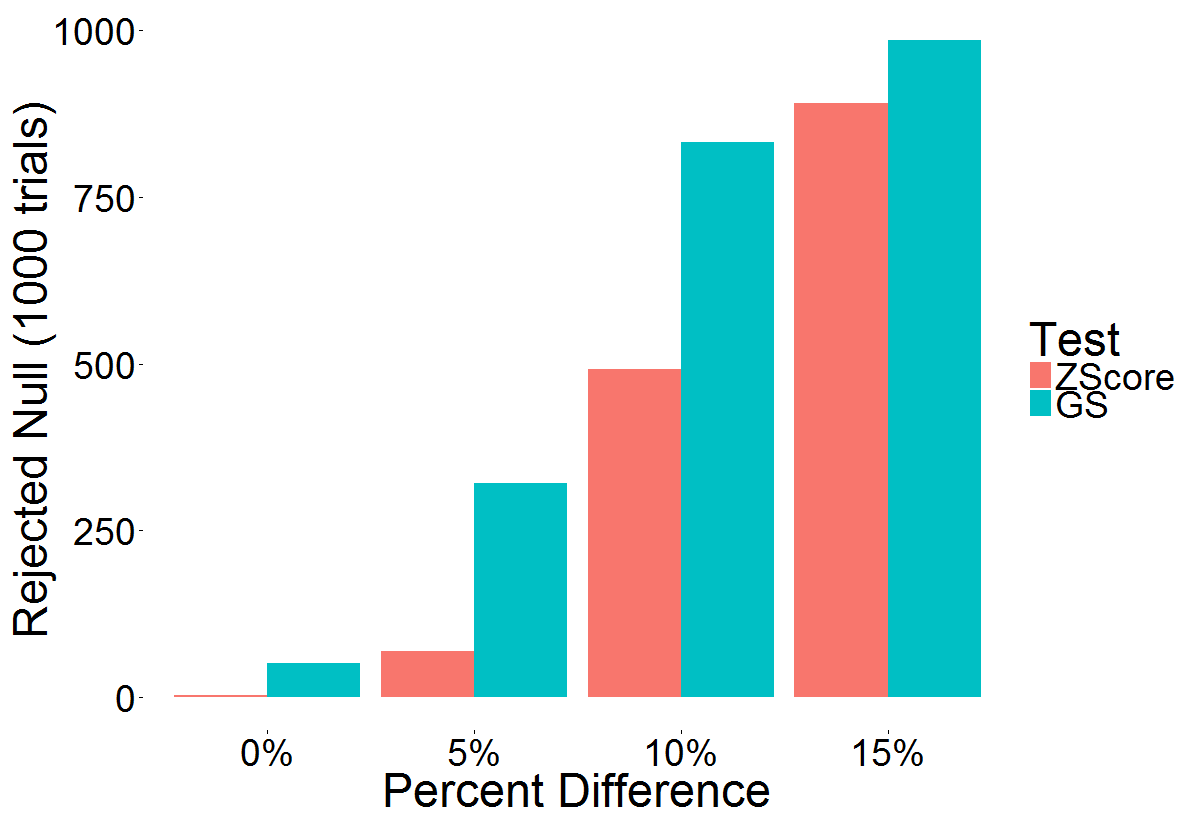}
		\caption{Sample size=1000}
		\label{subfigb}
	\end{subfigure}\hfill 
	\caption{Comparing Z-score test and GS test for correlated proportions, X-axis shows percent difference being tested and Y-axis is number of times Null hypothesis was rejected at $\alpha=0.05$ }
	\label{figabc}
\end{figure}

Ideally, we can compare $\hat{P_C}$ for different models. However,  major issue is the use of same dataset to build multiple models, which results in correlated precision values. Which if not accounted for, will result in biased inference. Another issue is the use of statistical tests designed to compare proportions, precision being a conditional probability is inherently different. When comparing precision or recall, it is not uncommon to use a Z-score test. Figures \ref{subfiga} and \ref{subfigb} show the reason to avoid such comparisons. For these plots, correlated values were generated using copula based simulations with a predefined difference. Z-score test being designed to compare independent proportions has lower power compared to GS test when there is a correlation.

It is also often overlooked that some classes might be of greater interest compared to others, thus requiring higher precision scores. Reporting and comparing average precision as in \ref{mainprec}  will dilute this effect. An example is of classifying a malignant tumor from others, where we would prefer a model with higher precision for malignant class. Second example can be of desired higher precision to identify stop signs and pedestrians for an autonomous vehicle. Classwise comparisons are also advantageous when sample size is large, i.e. when even small differences can be statistically significant, this is of a special concern in todays age of big data where problems involving deep learning can easily surpass a million observations.

As McNemar's test \citep{McNemar:1947}, paired T-test and Wilcoxon's signed rank test \citep{wilcoxon1945individual} can be used to compare classifiers based on their sensitivity, specificity and accuracy or mean of it. There is a need of appropriate statistical test that can be used to compare models using precision. As much time and effort goes into model building, model comparison should be treated the same. To best of our knowledge, there is no present literature in machine learning that reviews or introduces any tests to compare correlated precision values. However, similar studies are present for other metrics \citep{aslanstatistical,benavoli2014bayesian,demvsar2006statistical,dietterich1998approximate,joshi2002evaluating,nadeau2003inference}.

In this paper we present survey of statistical methods that can be used to compare classifiers based on precision. Comparisons are made on per-class basis, with methods provided to combine inference for an overall classifier comparison. Methods are introduced to compare classifiers in cross validation (single $k$-fold and $n$ times $k$-fold) settings commonly used by practitioners. We show that these methods can be used for simultaneous comparison of multiclass multiple classifiers.  We also present a partial Bayesian approach to update precision when class prevalence is known and demonstrate application of these methods to compare models based on deep architectures. We intend to enrich machine learning literature by providing methods to be used for model comparison using precision. Methods presented are not intended to replace or compete with existing statistically sound methods, but to supplement them. 

Next section presents an overview of statistical methods followed by empirical evaluation, our extensions and our recommendations and conclusion.

\section{Statistical tests}
This section introduces statistical tests that can be used for classifier comparison using precision alone. We start with standard notation and move on to introduce statistical tests based on marginal regression framework and relative precision.

\subsection{Standard notation}
We start by defining precision. Given two binary classifiers $C_1$ and $C_2$, we can write the results as Table \ref{t1} and Table \ref{t2}.
\begin{table}[h]
	\caption{True label vs predicted label for $C_1$.}
	\label{t1}
	\begin{center}
		\begin{small}
			\begin{sc}
				\begin{tabular}{lcccr}
					\multicolumn {5}{c}{$C_1$ Predicted label} \\ \hline
					&   & 1 & 0 & Total \\ \hline
					& 1 & \ a & \ b & \ $T_4$ \\
					True label & 0 & \ c & \ d & \ $T_3$  \\ \hline
					& Total & $T_1$ & $T_2$\\ \hline
				\end{tabular}
			\end{sc}
		\end{small}
	\end{center}
\end{table}

\begin{table}[h]
	\caption{True label vs predicted label for $C_2$.}
	\label{t2}
	\begin{center}
		\begin{small}
			\begin{sc}
				\begin{tabular}{lcccr}
					\multicolumn {5}{c}{$C_2$ Predicted label} \\ \hline
					&   & 1 & 0 & Total \\ \hline
					& 1 & \ e & \ f & \ $T_8$ \\
					True label & 0 & \ g & \ h & \ $T_7$  \\ \hline
					& Total & $T_5$ & $T_6$\\ \hline
				\end{tabular}
			\end{sc}
		\end{small}
	\end{center}
\end{table}
Estimated precision for classifier $C_1$ and $C_2$ can be written as
\begin{equation} \label{precision_define}
	\hat{{P}}_{C1}= \dfrac{A}{A+C}, \hat{{P}}_{C2}= \dfrac{E}{E+G}
\end{equation}

It is clear from \eqref{precision_define} that precision is the probability of making correct predictions conditioned on classifier predicted label, i.e. $Pr(True-label|Predicted-label)$.

Our null hypothesis for comparing $\hat{{P}}_{C1}$ and $\hat{{P}}_{C2}$ is
\begin{equation}
	H_0: \hat{{P}}_{C1}=\hat{{P}}_{C2}
\end{equation}
i.e. "For a given training set, the estimated precision for classifiers $C_1$ and $C_2$ is not statistically significantly different."

But as same dataset is used to calculate $\hat{{P}}_{C1}$ and $\hat{{P}}_{C2}$, values compared are not independent. Also being a conditional probability, precision does not lends itself well for methods designed for proportions. A suitable test that takes these factors into account is needed. Several such tests are presented in following sections.

\subsection{Methods based on marginal regression framework}
Let $D_i$ be true label where 1 is the class label of interest and 0 otherwise, $x_i$ is the predicted label by classifier and $f_i$ is the classifier used (1 for $C_1$ and 0 for $C_2$). We can define $P_{C1}$ and $P_{C2}$ as
\begin{equation}\label{class1}
	P_{C1} = Pr(D_i=1|f_i=1,x_i=1)
\end{equation}
\begin{equation}\label{class2}
	P_{C2} = Pr(D_i=1|f_i=0,x_i=1)
\end{equation}
i.e. probability of true label classified correctly conditioned on predicted label for classifier one and classifier two.

\cite{leisenring2000comparisons} used methods based on Generalized Estimating Equations (GEE) \citep{liang1986longitudinal} by restructuring the performance data of a medical diagnostic test to fit a marginal regression framework. We shall do the same, in our case, restructured data will have one row per classifier prediction. For a two classifier comparison we will have two rows per observation. Implementation details are given in Algorithm \ref{algo1}. A brief primer on marginal regression framework and GEE estimation in given in appendix.

As the dataset has repeated observations (multiple observations per data point), we use GEE based Generalized Linear Model (GLM) with a \emph{logit} link. Parameter estimation and associated standard errors are calculated using robust sandwich variance estimates \citep{huber,white,liang1986longitudinal}.

\begin{equation}\label{logistic model}
	Pr(D_i=1|f_i,x_i=1) = \dfrac{exp(\alpha+\beta f_i)}{1+exp(\alpha+\beta f_i)}
\end{equation}
\begin{algorithm}
	\caption{GEE based comparison using logistic regression}\label{algo1}
	\begin{algorithmic}[1]
		\Require \\
		k: Number of classes \\
		c: number of classifiers to be compared  \\
		y: outcome/target class \\
		$\hat{y}$: predicted class \\
		
	\State Generate \emph{id} variable for each observation\\
	\For{classifier in 1:c}
	\State Save prediction as a
	\State Save classifier name and y as b 
	\State Merge a and b horizontally 	
	\EndFor
		
	\State 	Stack c datasets vertically to generate a single dataset d 
	
	\For{i in 1:k}
			\State Subset dataset d with $\hat{y}=i$ 
			\State Fit a GEE based GLM with binomial link and independent working correlation matrix as shown in \eqref{logistic model} 
			\State Save required parameters 
	\EndFor
	\end{algorithmic}
\end{algorithm}

For a two classifier comparison $exp(\beta)$ is the ratio of true prediction given classifier predicted value from $C_2$ vs $C_1$. It describes the degree to which one classifier is more predictive of true classification than other.

Advantage of this method is the possibility of simultaneous comparison of multiple classifiers over multiple datasets. Model estimates such as odds ratio and related confidence intervals can be calculated for supplemental information.

\subsubsection{Empirical Wald test}
After fitting our GLM \eqref{logistic model}, Wald test for null hypothesis can be used to test $H_0: \hat{\beta} = 0$ .Which is equivalent to $H_0: \hat{P}_{C1}=\hat{P}_{C2}$, given as 
\begin{equation}
	\hat{T}_W = \dfrac{\hat{\beta^2}}{V(\hat{\beta})} 
\end{equation}
where denominator is second(last) diagonal element of empirical variance-covariance matrix.

Reformulation of empirical Wald test statistic was given in \cite{kosinski2013weighted} as
\begin{equation}\label{waldreform}
	\hat{T}_W = \dfrac{\hat{\beta^2}\times{\hat{P}}_{C1}(1-{\hat{P}}_{C1}){\hat{P}}_{C2}(1-{\hat{P}}_{C2})}{\dfrac{{\hat{P}}_{C1}(1-{\hat{P}}_{C1})}{T_1}+\dfrac{{\hat{P}}_{C2}(1-{\hat{P}}_{C2})}{T_5}-2{\hat{C}_P}(\dfrac{1}{T_1}+\dfrac{1}{T_5})}
\end{equation}
where $\hat{\beta}$ is the estimated regression parameter from \eqref{logistic model}, $T_i$ are totals from Table \ref{t1} and \ref{t2}. 

It is to be noted that GEE based Wald test statistic is similar to multinomial Wald statistic as $\hat{\beta} = logit\hat{{P}_{C1}} - logit\hat{{P}_{C2}}$. This relation is further expanded on in appendix.

\subsubsection{Score test}
Score test statistic based on GEE was given in \cite{leisenring2000comparisons} as
\begin{equation} \label{scoretest}
	\hat{T}_S = \dfrac{\{\sum_{i=1}^{N} D_i(T_i - m_i\bar{Z})\}^2}{\sum_{i=1}^{N}(D_i - \bar{D})^2(T_i - m_i\bar{Z})^2}
\end{equation}
where $T_i=\sum_{j=1}^{m_i}Z_{ij} $ is number of positive predicted labels for observation $i$. In a two classifier setting, $T_i$ is the indicator variable for correct predictions. Also
\begin{equation}
	\bar{D}=\dfrac{(\sum_{i=1}^{N}\sum_{j=1}^{m_i}D_{ij})}{\sum_{i=1}^{N}m_i}
\end{equation}
and
\begin{equation}
	\bar{Z} = \dfrac{(\sum_{i=1}^{N}\sum_{j=1}^{m_i}Z_{ij})}{\sum_{i=1}^{N}m_i}
\end{equation} 
$N$ is the number of observations with at least one true predicted label and $m_i$ is number of true predicted labels for $i^{th}$ observation.

Simple reformulation of complicated \eqref{scoretest} was given by \cite{kosinski2013weighted} as
\begin{equation}
	\hat{T}_S = \dfrac{({\hat{P}}_{C1} - {\hat{P}}_{C2})^2}{\{({\hat{P}}_{CP}(1 - {\hat{P}}_{CP})+\hat{W}_P - 2{\hat{C}_P}\}(\dfrac{1}{T_1}+\dfrac{1}{T_5})}
\end{equation}
where 
\begin{equation}
	\hat{W}_P = (2{\hat{P}}_{CP} - {\hat{P}}_{C1} - {\hat{P}}_{C2})(2{\hat{P}}_{CP} - 1)
\end{equation} 
and ${\hat{P}}_{CP}$ is pooled precision, estimated from Table 1 and 2 as
\begin{equation}
	\dfrac{A+E}{A+C+E+G}.
\end{equation} 

\subsection{Methods based on relative precision}
Table \ref{t1} and \ref{t2} can be restructured as Table \ref{t3}
\begin{table}[h]
	\caption{True label vs predicted label for $C_1$ and $C_2$}
	\label{t3}
	\begin{center}
		\begin{small}
			\begin{sc}
				\begin{tabular}{lllll}
					&True label=0      &      & True label=1  &      \\ 
					\hline
					& $C_2$=1 & $C_2$=0 & $C_2$=1 & $C_2$=0 \\ \hline
					$C_1$=1 & $n_1$   & $n_2$   & $n_5$   & $n_6$   \\
					$C_1$=0 & $n_3$   & $n_4$   & $n_7$   & $n_8$  \\ \hline
				\end{tabular}
			\end{sc}
		\end{small}
	\end{center}
\end{table}
and estimated precision for $C_1$ and $C_2$ is now rewritten as
\begin{equation}
	{\hat{P}}_{C1} = (n_5 + n_6)/(n_1+n_2+n_5+n_6)
\end{equation}
\begin{equation}
	\hat{{P}}_{C2} = (n_5 + n_7)/(n_1+n_3+n_5+n_7)
\end{equation}

Relative precision (RP) ${\hat{P}}_R$ is defined as 
\begin{equation}
	{\hat{P}}_R = \dfrac{{\hat{P}}_{C1}}{{\hat{P}}_{C2}}
\end{equation}
Using log transformation, variance of $log{\hat{P}}_R$ is estimated with $\hat{\sigma}^2_P/N$. Where
\begin{multline}
	\hat{\sigma}^2_P = \dfrac{1}{(n_5+n_7)(n_5+n_6)}\times \{n_6(1-{\hat{P}}_{C2})+n_5({\hat{P}}_{C2} - {\hat{P}}_{C1})\\
	+2(n_7+n_3){\hat{P}}_{C1}\times{\hat{P}}_{C2}+n_7(1-3{\hat{P}}_{C1})\}
\end{multline}
$100(1-\alpha)\%$ confidence intervals are then constructed as
\begin{equation}\label{logrelative}
	log\hat{P}_R \pm Z_{1-\alpha/2}\sqrt{\dfrac{\hat{\sigma_P}^2}{N}}
\end{equation}
\ref{logrelative} is exponentiated to obtain upper and lower limits of ${\hat{P}_R}$.
Confidence intervals from \eqref{logrelative} can be used to test 
\begin{equation}
{\hat{P_R}}=\xi
\end{equation}
where 
\begin{equation}
H_0:\hat{P}_{C1}=\hat{P}_{C2} \iff \xi=1 
\end{equation}
$H_0$ is rejected if lower confidence interval of ${\hat{P}_R}$ is greater than $\xi$ or upper confidence interval is less than $\xi$.

\section{Empirical evaluation}
\subsection{Experimental setup}
To demonstrate application feasibility of methods described in this paper, we used publically available datasets from UCI machine learning repository \citep{uci} with varying characteristics and sample sizes as shown in Table \ref{data}.
\begin{table}[h]
	\caption{Datasets used for evaluation}
	\label{data}
	\begin{center}
		\begin{small}
			\begin{sc}
				\begin{tabular}{lllll}
					\hline
				Dataset   & Instances & Attributes & Class \\
					\hline
				Wilt      & 4889      & 6          & 2     \\
				Diabetic Retinopathy  & 1151      & 20         & 2     \\
				Phishing  & 2456      & 30         & 2     \\
				Bank note & 1372      & 5          & 2     \\
				Magic     & 19020     & 11         & 2     \\
				Urban land cover      & 675       & 148        & 9   \\  \hline
				\end{tabular}
			\end{sc}
		\end{small}
	\end{center}
\end{table}

If a dataset was not already partitioned, training-test split of 70\%-30\% was used. Although most evaluations were performed using fixed training and test splits, same procedures can be adapted when using cross validation as shown in following sections.

Random forest\citep{breiman2001random} with 1000 trees and Naive Bayes were used for initial comparisons. All comparisons were implemented in R\citep{Rsoftware,halekoh2006r,stock2013dtcompair,hongying2014modified}. Sample code is made publically available (\url{https://github.com/lgondara/prec_compare}).

\subsection{Comparison using Generalized Score and Empirical Wald test}
Per-class precision was calculated for all datasets using random forest and Naive Bayes. Values were then compared using Generalized Score (GS) and Wald test statistic (GW). Results are shown in Table \ref{GSGWtable}.
\begin{table}[h]
	\caption{Comparison of $C_1$ and $C_2$ using GS and GW}
	\label{GSGWtable}
	\begin{center}
		\begin{small}
			\begin{sc}
				\begin{tabular}{llllll}
					\hline
					
					Dataset    & Class & NB   & RF   & P-GS            & P-GW            \\\hline
					
					Wilt       & N     & 0.65 & 0.74 & \textless0.0001 & \textless0.0001 \\
					& W     & 0.73 & 0.98 & 0.001           & 0.003           \\
					Diab. Ret. & 0     & 0.54 & 0.63 & 0.002           & 0.0002           \\
					& 1     & 0.76 & 0.67 & 0.07            & 0.09            \\
					Phishing   & -1    & 0.95 & 0.98 & \textless0.0001 & \textless0.0001 \\
					& 1     & 0.92 & 0.96 & \textless0.0001 & \textless0.0001 \\
					Bank note  & 0     & 0.83 & 0.99 & \textless0.0001 & 0.0001 \\
					& 1     & 0.85 & 0.98 & \textless0.0001 & \textless0.0001 \\
					MAGIC      & G     & 0.72 & 0.88 & \textless0.0001 & \textless0.0001 \\
					& H     & 0.70 & 0.87 & \textless0.0001 & \textless0.0001 \\
					Land Cover & Asp  & 0.94 & 0.95 & 0.87   & 0.87   \\
					& Bld & 0.91 & 0.85 & 0.04   & 0.04  \\
					& Cr      & 0.77 & 0.69 & 0.25   & 0.26   \\
					& Cnr & 0.85 & 0.78 & 0.04   & 0.05   \\
					& Grs    & 0.76 & 0.75 & 0.77   & 0.77   \\
					& Pl     & 0.92 & 0.92 & 0.95   & 0.95   \\
					& Shd   & 0.82 & 0.79 & 0.48   & 0.48   \\
					& Sl     & 0.36 & 0.60 & 0.01   & 0.02   \\
					& Tr     & 0.70 & 0.86 & 0.0001 & 0.001 \\ \hline
				\end{tabular}
			\end{sc}
		\end{small}
	\end{center}
	NB: Naive Bayes (Precision), RF: Random forest (Precision), P-GS: P-value from GS, GW: P-value from GS.
\end{table}
Lower p-values (typically $<0.05$) would signify a statistically significant difference between precision values of two classifiers. Results from GS and GW statistics agree for all comparisons. GS has more power and performs better with small sample size compared to GW as was also noticed by \cite{leisenring2000comparisons}. 

\subsection{Comparison based on relative precision}
Concerns around use/misuse of p-values \citep{nhst1,nhst2,nhst3} can be alleviated by using relative Precision (RP) and related confidence intervals (CIs). Although if necessary, p-value and a test statistic can be calculated as well. Comparison results using RP are shown in Table \ref{RPtable}.

\begin{table}[h]
	\caption{Comparison of $C_1$ and $C_2$ using Relative precision}
	\label{RPtable}
	\begin{center}
		\begin{small}
			\begin{sc}
				\begin{tabular}{llll}
					\hline
					Dataset    & Class & RP(95\% CI)           & P-value               \\ \hline
					Wilt       & N     & 0.88 (0.85,0.92) & \textless0.0001 \\
					& W     & 0.75 (0.62,0.91) & 0.003           \\
					Diab. Ret. & 0     & 0.85 (0.78,0.92) & 0.0001          \\
					& 1     & 1.13 (0.99,1.29) & 0.06            \\
					Phishing   & -1    & 0.97 (0.96,0.98) & \textless0.0001 \\
					& 1     & 0.96 (0.95,0.97) & \textless0.0001 \\
					Bank note  & 0     & 0.83 (0.79,0.88) & \textless0.0001 \\
					& 1     & 0.87 (0.82,0.92) & \textless0.0001 \\
					MAGIC      & G     & 0.82 (0.81,0.83) & \textless0.0001 \\
					& H     & 0.80 (0.77,0.83) & \textless0.0001 \\
					Land Cover & Asp  & 0.99 (0.92,1.10) & 0.87   \\
					& Bls & 1.07 (1.0,1.15)  & 0.04   \\
					& Cr      & 1.12 (0.92,1.35) & 0.26   \\
					& Cnr & 1.1 (1.0,1.17)   & 0.05   \\
					& Grs    & 1.02 (0.90,1.16) & 0.77   \\
					& Pl    & 1.01 (0.81,1.3)  & 0.95   \\
					& Shd   & 1.03 (0.94,1.14) & 0.48   \\
					& Sl     & 0.6 (0.39,0.93)  & 0.02   \\
					& Tr     & 0.81 (0.73,0.91) & 0.0002 \\ \hline
				\end{tabular}
			\end{sc}
		\end{small}
	\end{center}
	RP: Relative precision, 95\% CI: Confidence intervals, comparisons are based on "$NB/RF$"
\end{table}
Results using RP are in agreement with GW and GS. If just using RP and related confidence intervals, standard statistical interpretation can be used. CIs not including '1' indicate a statistical significant difference.

\begin{figure}[h!]
	\centering
		\centerline{\includegraphics[scale=0.9]{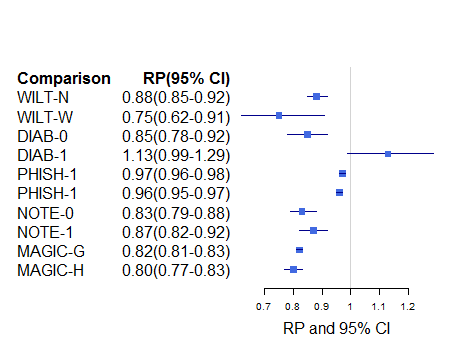}}
		\caption{Forest plot for relative comparison with 95\% confidence intervals}
		\label{forest}
\end{figure}

Another advantage of using RP is the nice graphical representation of results it lends itself to. An example of this is shown in Figure \ref{forest} with results plotted from Table \ref{RPtable} for first five datasets using a forest plot. Box represents point estimate with extended lines representing 95\% CIs. Reference line at '1' is plotted for visual inspection of a statistical significant difference. Confidence intervals not overlapping the reference line are considered significant.

\subsection{Combining inference}
Investigators are often interested in testing a global $H_0$ of an overall classifier comparison. Methods presented so far provide a per-class granular control, but an overall comparison is still desired. Results in Table \ref{GSGWtable} and Table \ref{RPtable} replicate a multiple comparison scenario, which if not accounted for can pose a challenge to control classifier wide Type I error rate. Common methods to adjust for multiple comparisons include Family Wise Error Rate (FWER) correction or controlling for False Discovery Rate (FDR). But, in this case we also need to acknowledge dependence between p-values, resulting from probable contribution of observations across classes. Hence, specially tailored methods to combine dependent p-values are needed.

For dependent p-values, the distribution of combined test statistic does not have an explicit analytical form. It is approximated using a scaled version \citep{li2011combined} with a new $\chi^2$ distribution. Satterthwaite method \citep{satter} was used by  \cite{hongying2014modified} to derive new degrees of freedom, scaled test statistic in this case is an extension of \cite{lancaster1961combination} and is given as
\begin{equation}
T_A=cT\approx\chi^2_\nu
\end{equation}
where
\begin{equation}
 c=\nu/E(T), \nu=2(E(T)^2/var(T))
\end{equation}
and
\begin{equation}
E(T)=\sum_{i=1}^{n} w_i
\end{equation} 
\begin{equation}
Var(T)=2 \sum_{i=1}^{n}w_i+2\sum_{i<j}p_{ij} 
\end{equation}
\begin{equation}
p_{ij}=cov(\gamma^{-1}_{(w_i/2,2)}(1-p_i),\gamma^{-1}_{(w_i/2,2)}(1-p_j))
\end{equation} 

$T$ is test statistic from \cite{lancaster1961combination} and $p_{ij}$ takes correlated p-values into account. When the covariance $p_{ij}$ is unknown, permutation or bootstrap methods can be used to simulate large enough sample (usually $\ge$ 1000) of p-values.

A much simpler method by Simes \citep{simes1986improved} can be used as well. For an ordered set of $L$ p-values, we have
\begin{equation}
Pr\left \{ \bigcup_{i}^{L}(p_{(i)}<i\alpha/L) \right \}<\alpha
\end{equation} 
rejecting global $H_0={H_1,...,H_L}$, if $p_{(i)}<i\alpha/L$ for at least one $i$. Global p-value is then given as $min\{L_{p(i)}/i\}$. Originally designed for independent p-values, it has been shown \citep{sarkar1997simes} to work well for positively correlated p-values.

To demonstrate, we focus on combining inference from one type of tests and on first five datasets where we would expect a combined test to reject global $H_0$. We used p-values from individual class comparisons using GS test. Method from \cite{hongying2014modified} was used to combine p-values as a positive correlation cannot be guaranteed. Table \ref{comb-p} shows the results. As expected, combined p-values are statistically significant at $\alpha<0.05$. Rejecting global $H_0$ in concordance with individual precision comparisons.

\begin{table}[h]
	\caption{Test of global null hypothesis using combined p-values}
	\label{comb-p}
	\begin{center}
		\begin{small}
			\begin{sc}
				\begin{tabular}{llll}
					\hline
					
					Dataset   & Class & P-value               & Combined P-value \\   
					\hline                   \\
					Wilt      & N     & \textless0.0001 & \multirow{2}{*}{\textless0.0001} \\
					& W     & 0.001           &                                  \\
					Diabetes  & 0     & 0.0002          & \multirow{2}{*}{0.002}           \\
					& 1     & 0.07            &                                  \\
					Phishing  & -1    & \textless0.0001 & \multirow{2}{*}{\textless0.0001} \\
					& 1     & \textless0.0001 &                                  \\
					Bank note & 0     & \textless0.0001 & \multirow{2}{*}{\textless0.0001} \\
					& 1     & \textless0.0001 &                                  \\
					MAGIC     & G     & \textless0.0001 & \multirow{2}{*}{\textless0.0001} \\
					& H     & \textless0.0001 &      \\ \hline                           
				\end{tabular}
			\end{sc}
		\end{small}
	\end{center}
\end{table}

\subsection{Multiple classifier comparison}
GEE based marginal regression framework can be used to compare multiple multiclass classifiers. With a logistic regression model, using state-of-the-art classifier as the reference category, we can compare the performance of new proposed models to state-of-the-art. This procedure is valuable in large scale testing where we want to compare tens of classifiers to select a best fit for a problem. We have used two datasets to show its feasibility. Random forest with 50 trees (RF2) and Support Vector Machines (SVM) were added, resulting in a four classifier comparison. Results are shown in Table \ref{multclass}. P-values $< \alpha$ inform us that at least one of the precision values is statistically significantly different from others. Magnitude and size of difference can be estimated from parameter estimates/odds ratio and related CIs.
\begin{table}[h]
	\caption{Multiple classifier comparison}
	\label{multclass}
	\begin{center}
		\begin{small}
			\begin{sc}
				\begin{tabular}{lllllll}
					\hline
					
					Dataset  & Class & NB   & RF1  & SVM  & RF2  & P-value   
					\\
					\hline
					
					Wilt     & N     & 0.65 & 0.74 & 0.68 & 0.74 & \textless0.0001 \\
					& W     & 0.73 & 0.98 & 1.0  & 0.95 & \textless0.0001 \\
					Diabetic & 0     & 0.54 & 0.63 & 0.63 & 0.62 & 0.003           \\
					& 1     & 0.76 & 0.67 & 0.72 & 0.66 & 0.12     \\ \hline      
				\end{tabular}
			\end{sc}
		\end{small}
	\end{center}
\end{table}

\begin{table}[h]
	\caption{Multiple classifier comparison, odds ratio and 95\% confidence intervals}
	\label{multclass2}
	\begin{center}
		\begin{small}
			\begin{sc}
				\begin{tabular}{llllll}
					\hline
					
					Dataset  & Class & Comparison & OR   & LCL  & UCL   
					\\
					\hline
					
					Wilt    & N & RF1 vs NB    & 1.52 & 1.34 & 1.73 \\
						    & N & SVM vs NB    & 1.12 & 1.03 & 1.22 \\
						    & N & RF2 vs NB    & 1.51 & 1.33 & 1.71 \\ \hline      
				\end{tabular}
			\end{sc}
		\end{small}
	\end{center}
	OR: odds ratio, LCL: Lower confidence limit, UCL: Upper confidence limit
\end{table}

Table \ref{multclass2} shows the comparison of four learning algorithms on a class of dataset Wilt using odds ratio and related 95\% confidence intervals. Only one class for one dataset is used for demonstration. Wealth of information provided by these estimates cannot be emphasized enough. Inferential statements such as "Random forest with 1000 trees has 52\% (95\% CI: 34\%, 73\%) higher chances of detecting non wilted trees compared to Naive Bayes" can be made. Odds ratio of greater than 1 confirms that model being compared is performing better than reference model.

\subsection{Partial Bayesian update of precision}
We introduce here a special case of Bayes law, updating precision when class prevalence is known. As with most datasets used, it is understood that they are sampled from a larger population. When a class prevalence in population is known, precision can be updated as following
\begin{equation}
P_{Bayes}=\frac{S_s \times P_v}{S_s \times P_v+(1-S_p)\times (1-P_v)}
\end{equation}
where $S_s$ is classifier sensitivity, $P_v$ is population prevalence and $S_p$ is classifier specificity. This update is well known in medical statistics. Precision can be significantly changed with change in class prevalence \citep{altman}. 

Classic example is from medical diagnostics where disease prevalence in a population is known and it can be used to update precision. Another example can be in object detection, as in indoor vs. outdoor images where prevalence of certain objects would be greater indoors (Tables, chairs, kettle etc.) and some outdoors (cars, buses, traffic signs etc.). 

This update can be applied to most scenarios. Even using a justifiable assumption should yield better population level estimates compared to a non-informative approach. An example is shown in Table \ref{bayes-prec} where we have used diabetic retinopathy dataset to calculate precision. Then it is updated using population prevalence from \citep{schneider,lee}. Relative precision is the recommended method to compare updated precision values. Confidence intervals are suggested to be calculated using bootstrap methods. This update can still be used if prevalence is not known by substituting a normalized prevalence rate.

\begin{table}[h]
	\caption{Updating Precision using class prevalence}
	\label{bayes-prec}
	\begin{center}
		\begin{small}
			\begin{sc}
				\begin{tabular}{llll}
					\hline
					
					Method   & Class & $P_{old}$ & $P_{update}$  \\   
					\hline                   \\
					NB  & 0     & 0.54   & 0.35            \\
					RF  & 0     & 0.63   & 0.45      \\
					NB  & 1     & 0.76   & 0.87            \\
					RF  & 1     & 0.67   & 0.81      \\
					 \hline                           
				\end{tabular}
			\end{sc}
		\end{small}
	\end{center}
	$P_{old}$ is empirical precision value and $P_{update}$ is updated precision based on prevalence
\end{table}

\subsection{Comparison using Cross Validation}
Methods described in this paper have been only applied in fixed train-test split. In this section we show their applicability when using cross validation. We used GEE based GLM on $k$-fold and $n$ times repeated $k$-fold cross validation. For $k$-fold CV, we used a value of $k=10$. For $n$ times repeated $k$-fold CV, a value of 10 was used for $n$ keeping $k$ fixed at 10. After saving predictions for each fold and for each classifier, datasets are vertically stacked to generate a single dataset with multiple observations per record. Then a GEE based GLM is fitted. This is a slight modification to Algorithm \ref{algo1}, presented as Algorithm \ref{algo2}. Results are reported in Table \ref{10foldcv} using diabetic retinopathy dataset. Results from both cross validation variations agree with results using a fixed train-test split, albeit cross validation based statistical comparisons have more power in limited sample size setting. Same methods can be used for any resampling method used during CV.

\begin{table}[h]
	\caption{Results from 10 fold CV and 10 $\times$ 10 fold CV}
	\label{10foldcv}
	\begin{center}
		\begin{small}
			\begin{sc}
				\begin{tabular}{lll}
					\hline
					
					Class & 10 fold CV   & 10 $\times$ 10 fold CV  \\   
					\hline                   \\
					Disease  & 50.6 ($<$0.0001)     & 50.6 ($<$0.0001) \\
					Non-Disease  & 10.2 (0.001)     & 10.2 (0.001)  \\
					\hline                           
				\end{tabular}
			\end{sc}
		\end{small}
	\end{center}
	Numbers outside parenthesis are test statistic with p-values inside
\end{table}

\begin{algorithm}
	\caption{GEE based comparison using CV and logistic regression}\label{algo2}
	\begin{algorithmic}[1]
		\Require \\
		k: Number of classes \\
		c: number of classifiers to be compared  \\
		y: outcome/target class \\
		$\hat{y}$: predicted class \\
		f: number of folds \\
	
		\State Generate \emph{id} variable for each observation\\
		\For{classifier in 1:c}
			\While{folds $<$ f}
				\State Save prediction as a
				\State Save classifier name and y as b 
				\State Merge a and b horizontally 
			\EndWhile
			\State Stack f datasets vertically
		\EndFor
		
		\State 	Stack c datasets vertically to generate a single dataset d 
		
		\For{i in 1:k}
		\State Subset dataset d with $\hat{y}=i$ 
		\State Fit a GEE based GLM with binomial link and independent working correlation matrix
		\State Save required parameters 
		\EndFor
	\end{algorithmic}
\end{algorithm}

\subsection{Application to deep architectures}
With shift to deep learning, aided by availability of better hardware and larger datasets. We are at a point where models are trained and tested on tens of thousands of objects. This large sample size makes even small differences statistically significant. Also, as new models are proposed often claiming to perform better than state-of-the-art, thorough comparisons are vital. Methods presented in this paper, applied on a per class basis can provide additional insights and can overcome some of the issues. This section shows the application of precision based comparison to deep architectures. We use two modified versions of deep convolution network described in \cite{verydeep}. For a simple demonstration we use the models to classify images of cats and dogs from Kaggle dataset (cats vs dogs) \citep{kaggle}. As the original model was trained on 1000 image classes including many instances of cats and dogs \citep{imagenet},it has been modified to work as a binary classifier \citep{catsdogs}. Two versions used differed only on dropout rate, first had a dropout rate of 0.5 and second 0.7.  Precision outcomes for both classes from both models are reported in Table \ref{deeptable}. As expected, overall accuracy of both models is very similar (90.76 and 90.88 respectively). But, the difference can be clearly seen in class breakdown where both models perform better on different classes. This type of analysis can also be used to adjust hyper-parameters for optimal performance.

\begin{table}[h]
	\caption{Comparing deep architectures using precision}
	\label{deeptable}
	\begin{center}
		\begin{small}
			\begin{sc}
				\begin{tabular}{llll}
					\hline
					
					Class &Model 1   & Model 2 & P-value  \\   
					\hline                   \\
					dogs  & 0.926     & 0.916   & 0.17            \\
					cats  & 0.889     & 0.902   & 0.06      \\
					\hline                           
				\end{tabular}
			\end{sc}
		\end{small}
	\end{center}
	Values shown in table above are precision values compared using GS statistic
\end{table}

Although above described scenario is overly simplistic, it is to demonstrate the usefulness of presented methods to compare state-of-the-art. More complicated comparisons such as multiple object detection/classification in images can be implemented with similar ease.

\section{Replicability}
High replicability of a test statistic is vital, it does not only facilitate reproducible research but is also an estimate of the degree to which random partitioning and other dataset features are related to test results. We focus on replicability as a function of test dataset proportion and an overall sample size. 

We use replicability measure introduced by Bouckaert \citep{bouckaert2004evaluating,bouckaert2004estimating} based on number of rejections of $H_0$ , given as

\begin{equation}
	R(e)=\sum_{1\le i<j\le n}\frac{I(e_i=e_j)}{n(n-1)/2}
\end{equation}

where $e_i$ is outcome of $i$-th experiment, $n$ is total number of experiments. $e_i$ is 1 if $H_0$ is accepted and 0 otherwise. $I$ is an indicator function which is 1 when argument is true and 0 otherwise. Above statistic can also be calculated with a simpler formula

\begin{equation}
	R(e)=p(p-1)+q(q-1)/n(n-1)
\end{equation}
where $p$ is the number when $H_0$ is accepted and $q$ is when it is rejected. 

\begin{figure}[h!]
	\vskip 0.2in
	\begin{center}
		\centerline{\includegraphics[scale=0.4]{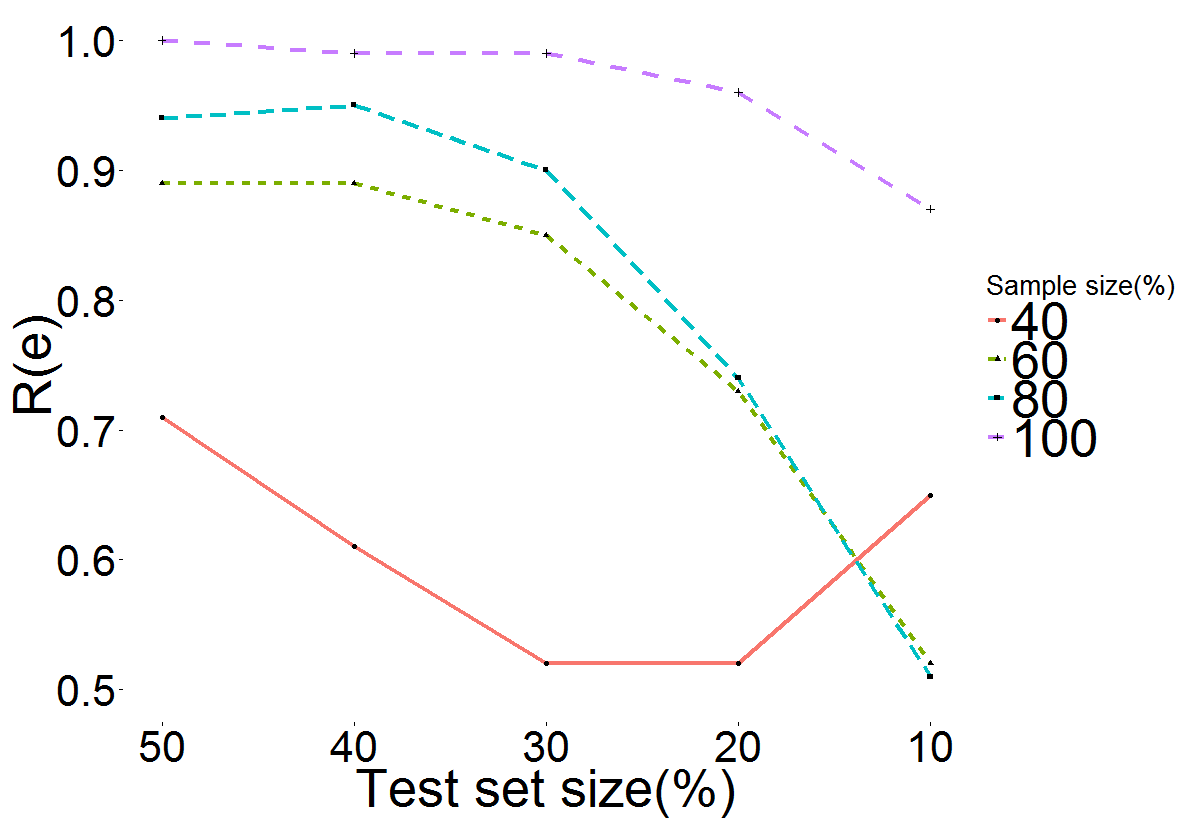}}
		\caption{Reproducibility as a function of overall sample size and test set proportion for Diabetic retinopathy data(initial p-value: 0.0002)}
		\label{plot1}
	\end{center}
	\vskip -0.2in
\end{figure} 

We used GS statistic and an initial $n$ of 600 from "Diabetic retinopathy" dataset with varying proportion of test set and initial sample sizes. Class "0" was used for this test. The dataset and class were chosen as the p-value was not too small or too large to skew replicability outcomes as highly significant or non-significant p-values will tend to be more replicable than marginally significant ones.

Results are shown in Figure \ref{plot1}. It can be seen that using full dataset of $n=600$, high replicability is maintained even when test set size is just 20\%. Replicability deteriorates with decreasing sample size. It is to be noticed that using 40\% of original dataset, replicability falls sharply and then increases with decreasing test set proportion. This is due to increased number of failures to reject $H_0$, which boosts replicability but with an opposite sign.

\section{Conclusion and Recommendations}
While machine learning literature is rich with evaluations and recommendations for statistical tests to compare classifiers based on classification accuracy, AUC, F-measure etc. It lacked a detailed study of statistical tests that can be used to compare classifiers based on precision or recall alone. Which are important performance metrics, especially for rare event classifiers. In this paper we have reviewed statistical methods based on marginal regression framework and Relative Precision. These can be used for classifier comparison using correlated precision values. We have presented empirical evaluation and implementation feasibility of these methods. As precision is usually calculated per-class,methods are presented to combine p-values for an overall classifier comparison. When a class prevalence is known, partial Bayesian update to precision is introduced. We have shown that the methods can be used in a cross validation setting and their application to compare deep architectures.

We recommend using GS statistic or RP for comparing two classifiers. Users concerned about use/misuse of p-values in statistical tests should use RP as results can be solely based on RP value and CIs. To simultaneously compare multiple classifiers, we recommend using GLM with GEE. Dai's method is recommended for combining dependent p-values over Simes' method as it retains appropriate power even when p-values are not positively correlated. Whenever possible, it is also recommended to use updated precision based on population prevalence.

\appendix
\section{Multinomial Wald test is similar to GEE based empirical Wald test}
Multinomial Wald statistic for comparing precision with \emph{logit} transformed values is given assuming cells in Table \ref{t1} and \ref{t2} to be multinomially distributed. Using the delta method, Wald statistic for testing $H_0 : \hat{{P}}_{C1} = \hat{{P}}_{C2}$ is given as

\begin{equation}\label{multwald}
	T_M = \dfrac{\{f(\hat{{P}_{C1}}) - f(\hat{{P}_{C2}})\}^2}{var\{\{f(\hat{{P}_{C1}}) - f(\hat{{P}_{C2}})\}\}}
\end{equation}

For $f(x)=x$ and estimated covariance matrix of $f(\hat{{P}_{C1}})$ and  $f(\hat{{P}_{C2}})$, we can rewrite \eqref{multwald} as

\begin{equation}\label{simplewald}
	\hat{T_M} = \dfrac{(\hat{{P}}_{C1} - \hat{{P}}_{C2})^2}{\dfrac{\hat{{P}}_{C1}(1-\hat{{P}}_{C1})}{T_1}+\dfrac{\hat{{P}}_{C2}(1-\hat{{P}}_{C2})}{T_5}-2{\hat{C_P}}(\dfrac{1}{T_1}+\dfrac{1}{T_5})}
\end{equation}

where

\begin{equation}
	{\hat{C_P}} = \dfrac{x(1-\hat{{P}}_{C1})(1-\hat{{P}}_{C2})+y\hat{{P}}_{C1}\hat{{P}}_{C2}}{T_1+T_5}
\end{equation}

and $x$ is number of times both classifiers predicted correct (positive concordance) and $y$ when both were wrong (negative concordance). We refer readers to \cite{kosinski2013weighted} for derivation of covariance matrix.

for $f(x)=logit(x)$, we have

\begin{equation}\label{logisticwald}
	T_{\hat{l_tM}} = \dfrac{(\hat{logit{P}}_{C1} - \hat{logit{P}}_{C2})^2 \times \hat{{P}}_{C1}(1-{\hat{P}}_{C1}){\hat{P}}_{C2}(1-{\hat{P}}_{C2})}{\dfrac{{\hat{P}}_{C1}(1-{\hat{P}}_{C1})}{T_1}+\dfrac{{\hat{P}}_{C2}(1-{\hat{P}}_{C2})}{T_5}-2{\hat{C_P}}(\dfrac{1}{T_1}+\dfrac{1}{T_5})}
\end{equation}.

Equivalency of \ref{logisticwald} and \ref{waldreform} can be seen when substituting

\begin{equation}
    \hat{\beta} = logit\hat{{P}_{C1}} - logit\hat{{P}_{C2}}
\end{equation}

\section{Generalized Estimating Equations (GEE)}
If we have dependent sampling, i.e. repeated measures, matched pairs etc., we need specialized modelling approaches to account for correlation as simpler models work on an assumption of independent response. In our case this dependence arises when we build multiple models on same dataset. To model our data to test for differences in precision values using Generalized Linear Models(GLM) or more specific \emph{logistic regression}, we will use Generalized Estimating Equations, which is an extension of GLMs, or specifically an estimating framework when responses are not independent. Regression models estimated using GEE are often referred to as \emph{marginal regression models} or \emph{population averaged models} i.e. inferences are made about population averages and term "marginal" signifies that mean response is modelled conditional only on covariates and not on other responses or random effects.

Basic idea of GEE is to model the mean response treating within observation correlation structure as a nuisance parameter. In this framework, we don't need to correctly specify correlation structure to get reasonable estimates for parameter coefficients and standard errors (both needed to calculate p-values for comparison and to get magnitude of difference in our case). Main difference between a GLM in independent observations scenario and a GEE based GLM is the need of modelling covariance structure of correlated responses. Then the model is estimated using quasi-likelihood rather than maximum likelihood.

Quasi-likelihood estimators are estimates of quasi-likelihood equations known as \emph{Generalized Estimating Equations}. There is no closed form solution in general, estimation is done using an iterative process. Standard errors can be calculated using the sandwich estimator, given as
\begin{equation}
    [X^T\hat{W}X]^{-1} [\sum_i X_i^T(y_i - \hat{\mu_i})(y_i - \hat{\mu_i})^TX_i]  [X^T\hat{W}X]^{-1}
\end{equation}

Full details of GEE or GLMs are out of scope for this paper. We refer readers to \citep{agresti2011categorical} for further details. However complex it may sound, it is implemented out of the box in all statistical packages.

\vskip 0.2in
\bibliography{example_paper}

\begin{thebibliography}{37}
\providecommand{\natexlab}[1]{#1}
\providecommand{\url}[1]{\texttt{#1}}
\expandafter\ifx\csname urlstyle\endcsname\relax
  \providecommand{\doi}[1]{doi: #1}\else
  \providecommand{\doi}{doi: \begingroup \urlstyle{rm}\Url}\fi

\bibitem[Agresti and Kateri(2011)]{agresti2011categorical}
Alan Agresti and Maria Kateri.
\newblock \emph{Categorical data analysis}.
\newblock Springer, 2011.

\bibitem[Altman and Bland(1994)]{altman}
Douglas~G Altman and J~Martin Bland.
\newblock Statistics notes: Diagnostic tests 2: predictive values.
\newblock \emph{Bmj}, 309\penalty0 (6947):\penalty0 102, 1994.

\bibitem[Anderson et~al.(2000)Anderson, Burnham, and Thompson]{nhst3}
David~R Anderson, Kenneth~P Burnham, and William~L Thompson.
\newblock Null hypothesis testing: problems, prevalence, and an alternative.
\newblock \emph{The journal of wildlife management}, pages 912--923, 2000.

\bibitem[Aslan et~al.()Aslan, Y{\i}ld{\i}z, and Alpayd{\i}n]{aslanstatistical}
Ozlem Aslan, Olcay~Taner Y{\i}ld{\i}z, and Ethem Alpayd{\i}n.
\newblock Statistical comparison of classifiers using area under the roc curve.

\bibitem[Benavoli et~al.(2014)Benavoli, Corani, Mangili, Zaffalon, and
  Ruggeri]{benavoli2014bayesian}
Alessio Benavoli, Giorgio Corani, Francesca Mangili, Marco Zaffalon, and
  Fabrizio Ruggeri.
\newblock A bayesian wilcoxon signed-rank test based on the dirichlet process.
\newblock In \emph{Proceedings of the 31st International Conference on Machine
  Learning (ICML-14)}, pages 1026--1034, 2014.

\bibitem[Blake and Merz(1998)]{uci}
Catherine Blake and Christopher~J Merz.
\newblock $\{$UCI$\}$ repository of machine learning databases.
\newblock 1998.

\bibitem[Bouckaert(2004)]{bouckaert2004estimating}
Remco~R Bouckaert.
\newblock Estimating replicability of classifier learning experiments.
\newblock In \emph{Proceedings of the twenty-first international conference on
  Machine learning}, page~15. ACM, 2004.

\bibitem[Bouckaert and Frank(2004)]{bouckaert2004evaluating}
Remco~R Bouckaert and Eibe Frank.
\newblock Evaluating the replicability of significance tests for comparing
  learning algorithms.
\newblock In \emph{Pacific-Asia Conference on Knowledge Discovery and Data
  Mining}, pages 3--12. Springer, 2004.

\bibitem[Breiman(2001)]{breiman2001random}
Leo Breiman.
\newblock Random forests.
\newblock \emph{Machine learning}, 45\penalty0 (1):\penalty0 5--32, 2001.

\bibitem[Chollet()]{catsdogs}
Francois Chollet.
\newblock Building powerful image classification models using very little data.
\newblock
  \url{(https://blog.keras.io/building-powerful-image-classification-models-using-very-little-data.html}.
\newblock Accessed: 2016-07-29.

\bibitem[Dem{\v{s}}ar(2006)]{demvsar2006statistical}
Janez Dem{\v{s}}ar.
\newblock Statistical comparisons of classifiers over multiple data sets.
\newblock \emph{The Journal of Machine Learning Research}, 7:\penalty0 1--30,
  2006.

\bibitem[Dietterich(1998)]{dietterich1998approximate}
Thomas~G Dietterich.
\newblock Approximate statistical tests for comparing supervised classification
  learning algorithms.
\newblock \emph{Neural computation}, 10\penalty0 (7):\penalty0 1895--1923,
  1998.

\bibitem[Gill(1999)]{nhst2}
Jeff Gill.
\newblock The insignificance of null hypothesis significance testing.
\newblock \emph{Political Research Quarterly}, 52\penalty0 (3):\penalty0
  647--674, 1999.

\bibitem[Halekoh et~al.(2006)Halekoh, H{\o}jsgaard, and Yan]{halekoh2006r}
Ulrich Halekoh, S{\o}ren H{\o}jsgaard, and Jun Yan.
\newblock s.
\newblock \emph{Journal of Statistical Software}, 15\penalty0 (2):\penalty0
  1--11, 2006.

\bibitem[Hongying~Dai and Cui(2014)]{hongying2014modified}
J~Hongying~Dai and Yuehua Cui.
\newblock A modified generalized fisher method for combining probabilities from
  dependent tests.
\newblock \emph{Frontiers in genetics}, 5, 2014.

\bibitem[Huber(1967)]{huber}
Peter~J Huber.
\newblock The behavior of maximum likelihood estimates under nonstandard
  conditions.
\newblock In \emph{Proceedings of the fifth Berkeley symposium on mathematical
  statistics and probability}, volume~1, pages 221--233, 1967.

\bibitem[Joshi(2002)]{joshi2002evaluating}
Mahesh~V Joshi.
\newblock On evaluating performance of classifiers for rare classes.
\newblock In \emph{Data Mining, 2002. ICDM 2003. Proceedings. 2002 IEEE
  International Conference on}, pages 641--644. IEEE, 2002.

\bibitem[Kaggle()]{kaggle}
Kaggle.
\newblock Kaggle dogs vs cats competitiion.
\newblock \url{(https://www.kaggle.com/c/dogs-vs-cats}.
\newblock Accessed: 2016-07-29.

\bibitem[Kosinski(2013)]{kosinski2013weighted}
Andrzej~S Kosinski.
\newblock A weighted generalized score statistic for comparison of predictive
  values of diagnostic tests.
\newblock \emph{Statistics in medicine}, 32\penalty0 (6):\penalty0 964--977,
  2013.

\bibitem[Lancaster(1961)]{lancaster1961combination}
HO~Lancaster.
\newblock The combination of probabilities: an application of orthonormal
  functions.
\newblock \emph{Australian Journal of Statistics}, 3\penalty0 (1):\penalty0
  20--33, 1961.

\bibitem[Lee et~al.(2015)Lee, Wong, and Sabanayagam]{lee}
Ryan Lee, Tien~Y Wong, and Charumathi Sabanayagam.
\newblock Epidemiology of diabetic retinopathy, diabetic macular edema and
  related vision loss.
\newblock \emph{Eye and Vision}, 2\penalty0 (1):\penalty0 1, 2015.

\bibitem[Leisenring et~al.(2000)Leisenring, Alono, and
  Pepe]{leisenring2000comparisons}
Wendy Leisenring, Todd Alono, and Margaret~Sullivan Pepe.
\newblock Comparisons of predictive values of binary medical diagnostic tests
  for paired designs.
\newblock \emph{Biometrics}, 56\penalty0 (2):\penalty0 345--351, 2000.

\bibitem[Li et~al.(2011)Li, Williams, and Cui]{li2011combined}
Shaoyu Li, Barry~L Williams, and Yuehua Cui.
\newblock A combined p-value approach to infer pathway regulations in eqtl
  mapping.
\newblock \emph{Statistics and Its Interface}, 4:\penalty0 389--401, 2011.

\bibitem[Liang and Zeger(1986)]{liang1986longitudinal}
Kung-Yee Liang and Scott~L Zeger.
\newblock Longitudinal data analysis using generalized linear models.
\newblock \emph{Biometrika}, pages 13--22, 1986.

\bibitem[McNemar(1947)]{McNemar:1947}
Quinn McNemar.
\newblock Note on the sampling error of the difference between correlated
  proportions or percentages.
\newblock \emph{Psychometrika}, 12\penalty0 (2):\penalty0 153--157, 1947.

\bibitem[Nadeau and Bengio(2003)]{nadeau2003inference}
Claude Nadeau and Yoshua Bengio.
\newblock Inference for the generalization error.
\newblock \emph{Machine Learning}, 52\penalty0 (3):\penalty0 239--281, 2003.

\bibitem[Nickerson(2000)]{nhst1}
Raymond~S Nickerson.
\newblock Null hypothesis significance testing: a review of an old and
  continuing controversy.
\newblock \emph{Psychological methods}, 5\penalty0 (2):\penalty0 241, 2000.

\bibitem[{R Core Team}(2015)]{Rsoftware}
{R Core Team}.
\newblock \emph{R: A Language and Environment for Statistical Computing}.
\newblock R Foundation for Statistical Computing, Vienna, Austria, 2015.
\newblock URL \url{https://www.R-project.org}.

\bibitem[Russakovsky et~al.(2015)Russakovsky, Deng, Su, Krause, Satheesh, Ma,
  Huang, Karpathy, Khosla, Bernstein, et~al.]{imagenet}
Olga Russakovsky, Jia Deng, Hao Su, Jonathan Krause, Sanjeev Satheesh, Sean Ma,
  Zhiheng Huang, Andrej Karpathy, Aditya Khosla, Michael Bernstein, et~al.
\newblock Imagenet large scale visual recognition challenge.
\newblock \emph{International Journal of Computer Vision}, 115\penalty0
  (3):\penalty0 211--252, 2015.

\bibitem[Sarkar and Chang(1997)]{sarkar1997simes}
Sanat~K Sarkar and Chung-Kuei Chang.
\newblock The simes method for multiple hypothesis testing with positively
  dependent test statistics.
\newblock \emph{Journal of the American Statistical Association}, 92\penalty0
  (440):\penalty0 1601--1608, 1997.

\bibitem[Satterthwaite(1946)]{satter}
Franklin~E Satterthwaite.
\newblock An approximate distribution of estimates of variance components.
\newblock \emph{Biometrics bulletin}, 2\penalty0 (6):\penalty0 110--114, 1946.

\bibitem[Schneider and S{\"u}veges(2004)]{schneider}
Mikl{\'o}s Schneider and Ildik{\'o} S{\"u}veges.
\newblock Retinopathia diabetica: magyarorsz{\'a}gi epidemiol{\'o}giai adatok.
\newblock \emph{Szem{\'e}szet}, 141:\penalty0 441--444, 2004.

\bibitem[Simes(1986)]{simes1986improved}
R~John Simes.
\newblock An improved bonferroni procedure for multiple tests of significance.
\newblock \emph{Biometrika}, 73\penalty0 (3):\penalty0 751--754, 1986.

\bibitem[Simonyan and Zisserman(2014)]{verydeep}
Karen Simonyan and Andrew Zisserman.
\newblock Very deep convolutional networks for large-scale image recognition.
\newblock \emph{arXiv preprint arXiv:1409.1556}, 2014.

\bibitem[Stock and Hielscher(2013)]{stock2013dtcompair}
C~Stock and T~Hielscher.
\newblock Dtcompair: comparison of binary diagnostic tests in a paired study
  design.
\newblock \emph{R package version}, 1, 2013.

\bibitem[White(1980)]{white}
Halbert White.
\newblock A heteroskedasticity-consistent covariance matrix estimator and a
  direct test for heteroskedasticity.
\newblock \emph{Econometrica: Journal of the Econometric Society}, pages
  817--838, 1980.

\bibitem[Wilcoxon(1945)]{wilcoxon1945individual}
Frank Wilcoxon.
\newblock Individual comparisons by ranking methods.
\newblock \emph{Biometrics bulletin}, 1\penalty0 (6):\penalty0 80--83, 1945.

\end{thebibliography}

\end{document}